# Model-free Optical Processors using *In Situ* Reinforcement Learning with Proximal Policy Optimization

Yuhang Li[1,2], Shiqi Chen[1,2], Tingyu Gong[3], and Aydogan Ozcan[*,1,2,4]

[1]Electrical and Computer Engineering Department, University of California, Los Angeles, CA, 90095, USA.

[2]California NanoSystems Institute (CNSI), University of California, Los Angeles, CA, USA.

[3]Computer Science Department, University of California, Los Angeles, 90095, USA.

[4]Bioengineering Department, University of California, Los Angeles, 90095, USA.

[*]Correspondence: Aydogan Ozcan. Email: ozcan@ucla.edu

## Abstract

Optical computing holds promise for high-speed, energy-efficient information processing, with diffractive optical networks emerging as a flexible platform for implementing task-specific transformations. A challenge, however, is the effective optimization and alignment of the diffractive layers, which is hindered by the difficulty of accurately modeling physical systems with their inherent hardware imperfections, noise, and misalignments. While existing *in situ* optimization methods offer the advantage of direct training on the physical system without explicit system modeling, they are often limited by slow convergence and unstable performance due to inefficient use of limited measurement data. Here, we introduce a model-free reinforcement learning approach utilizing Proximal Policy Optimization (PPO) for the *in situ* training of diffractive optical processors. PPO efficiently reuses *in situ* measurement data and constrains policy updates to ensure more stable and faster convergence. We experimentally validated our method across a range of *in situ* learning tasks, including targeted energy focusing through a random diffuser, holographic image generation, aberration correction, and optical image classification, demonstrating in each task better convergence and performance. Our strategy operates directly on the physical system and naturally accounts for unknown real-world imperfections, eliminating the need for prior system knowledge or modeling. By enabling faster and more accurate training under realistic experimental constraints, this *in situ* reinforcement learning approach could offer a scalable framework for various optical and physical systems governed by complex, feedback-driven dynamics.

# Introduction

As the demand for faster, more efficient artificial intelligence (AI) computation grows, physical neural networks (PNNs), which perform computation using analog systems, have emerged as an alternative to traditional digital processors. By leveraging physical processes such as light propagation, electrical response, or acoustic vibration, PNNs offer the potential for ultra-low-latency, energy-efficient inference and edge computing[1–4]. This paradigm is especially promising in the domain of optical computing, where the input information is rapidly processed using thin optical components[5]. A variety of optical platforms have been explored, including diffractive optical networks[6–9], integrated photonic neural networks[10–13], and optical reservoir computing[14–17], each offering unique advantages for accelerating light-based computation.

Designing optical computing systems typically involves a two-step process: first, digitally emulating and optimizing the physical parameters of the hardware in a simulated environment; and second, deploying the optimized configuration into a real-world physical system[6,13,18–20]. *In silico* training methods rely on physics-based forward models or neural network surrogates to construct digital twins of PNNs, which are then optimized *in silico* for specific tasks, as shown in **Fig. 1a**. However, this simulation-driven *in silico* approach faces inherent limitations. Accurately modeling the physical system is often undermined by the simulation-to-reality gap[21–23]; real-world systems are affected by noise, optical misalignments, and fabrication or device imperfections that are difficult to accurately model or know a priori[24]. Even when a reasonable model exists, simulating physical processes requires fine discretization of space and time, making optimization computationally expensive and susceptible to numerical errors[25,26]. Some of the recent approaches have explored a hybrid strategy by combining digital and physical elements during training. In these physics-aware training methods, a fixed physical system performs the forward pass, while a co-optimized digital model computes the backward pass via gradient-based updates[1,27]. Although hybrid strategies that combine physical feedback with a digital model relax the strict requirement of having a perfect digital twin, they still depend on maintaining a close correspondence between the digital and physical worlds, a challenge that grows with system complexity. If the system drifts or exhibits unmodeled/unknown dynamics or aberrations, the gradient estimates become unreliable.

To eliminate this dependence on accurate models, model-free training algorithms have recently gained attention[2,28–33]. In particular, *in situ* training methods—where the optimization is performed directly on the physical hardware—have begun to emerge. One set of approaches involves optical implementations of backpropagation, demonstrated in both linear and nonlinear optical networks, where gradients are computed using forward and backward light propagation. While powerful, these approaches demand precise optical alignments, which might limit their practical applications in diverse or noisy experimental settings[29,34]. More flexible model-free optimization techniques, such as gradient-free and evolutionary algorithms, have also been explored[35–38]. These include methods based on perturbation strategies—such as Simultaneous Perturbation Stochastic Approximation (SPSA)—and those based on sampling from distributions[32,37,39–42], including

Genetic Algorithms (GA), Evolutionary Strategies (ES), swarm optimization, and reinforcement learning (RL). Recently, score-function-based gradient estimators have also been used to train optical systems, leveraging policy gradient (PG) methods to directly optimize the system's response without an explicit model[28,43,44]; see **Fig. 1b**. These approaches interact directly with the physical system for the forward pass and use reward signals to guide updates—without requiring explicit gradient computation through the hardware. Although these efforts have shown promising results, a key bottleneck shared by all these *in situ* model-free approaches is the relatively high cost of physical measurements. Unlike digital computations, capturing each physical measurement is a slow, sequential process constrained by the speed of optical hardware, e.g., spatial light modulators (SLMs) and optical sensors. This makes data collection time-consuming. Moreover, standard methods typically discard collected samples after each update, resulting in inefficient data usage and unreliable gradient estimates. As a result, updating the policy parameters based on each new batch of measurements can make data inefficiency a critical issue. Furthermore, standard PG methods often suffer from unstable convergence, significantly compounding this data inefficiency problem by requiring more iterations to reach an optimal solution. These drawbacks are especially problematic in physical systems, where each policy update requires real-world measurements and is therefore time-consuming and resource-intensive.

Here, we introduce a model-free, *in situ* reinforcement learning framework based on Proximal Policy Optimization (PPO) for training optical processors. PPO improves optical training efficiency by enabling multiple digital updates per batch of captured optical measurements and ensures stable convergence by constraining policy changes, even under noisy experimental conditions[45]. We experimentally demonstrated the effectiveness of this *in situ* optical learning approach across a variety of tasks—including energy focusing through a random, unknown diffuser, holographic image generation, aberration correction, and optical image classification. Crucially, our PPO-based method achieved substantially faster convergence and improved final performance across all tasks, without relying on physical system models or explicit gradient calculations. We anticipate that this robust, data-efficient training strategy will significantly advance the practical deployment of optical processors and it offers a generalizable framework for optimizing complex, feedback-driven physical systems.

## Results

Policy gradient is a class of reinforcement learning algorithms that directly optimize a parameterized policy, i.e., a function that maps system states to actions, by adjusting its parameters to maximize an expected reward. These methods estimate gradients that guide the policy to iteratively increase the likelihood of actions yielding higher rewards, while decreasing the likelihood of actions with lower rewards. In our *in situ* optical learning framework, we utilized PPO to achieve stability, data efficiency, and enhanced performance across various optical tasks. As shown in **Fig. 2a**, PPO-based training strategy improves the *in situ* reinforcement learning process by enabling data reuse. In each training round, we sample $M$ phase profiles $\{\varphi_j\}_{j=1}^{M}$ from the current policy $\pi_\theta$, display them sequentially on the SLM, and record the resulting optical measurements, corresponding to a certain desired optical task. These measurements are then used to estimate the advantage function and compute the PPO loss[46]; see **Methods** for details. PPO performs multiple optimization steps using the same set of measured physical/experimental data. This significantly improves data efficiency—an essential benefit in time-constrained *in situ* optical experiments. To prevent these repeated updates from causing instability or divergence, we used a clipped surrogate objective that constrains how much the policy can change in each step. The PPO objective is defined as[45]:

$$J^{PPO}(\theta) = -\mathbb{E}_{\varphi \sim \pi_\theta}\left[\min\left(r(\varphi;\theta)A'(\varphi), \mathrm{clip}(r(\varphi;\theta), 1-\epsilon, 1+\epsilon)A'(\varphi)\right)\right] \quad (1)$$

where $r(\varphi;\theta) = \frac{\pi_\theta(\varphi)}{\pi_{\theta_{old}}(\varphi)}$ is the probability ratio between the current policy $\pi_\theta(\varphi)$ and previous policy $\pi_{\theta_{old}}(\varphi)$, and $A'(\varphi)$ denotes the normalized advantage function, which measures how much better or worse a sampled phase profile $\varphi$ performed compared to the average (see **Methods** for details). The clip operator constrains the probability ratio of $r(\varphi;\theta)$ to the range $[1-\epsilon, 1+\epsilon]$, where $\epsilon$ is a hyperparameter limiting the extent of the policy update to ensure stability. The core clipping mechanism prevents overly large updates by enforcing a conservative change in the policy distribution, thereby ensuring stable convergence even when using repeated samples of physical measurement data. We performed this surrogate loss optimization for $K$ iterations using the same batch of collected data before sampling new data. By reusing collected data and preventing overly large policy shifts, PPO offers a practical balance between exploration and exploitation, making it well-suited for data-constrained experimental scenarios.

We first validated our PPO-based model-free reinforcement learning strategy in a simulated optical classification task, where a diffractive layer was optimized to classify phase-encoded MNIST digits[47] (see **Methods** for details). **Figure 2b** quantitatively compares the convergence of PG and PPO. The PPO strategy using a single diffractive layer achieved a final test accuracy of ~80% and 3.2 times faster than PG. This accelerated convergence demonstrates PPO's effectiveness in optimizing complex optical transformations through reinforcement learning. Furthermore, **Fig. 2c** visualizes the evolution of the learned phase patterns over training iterations, showing that PPO rapidly learned clear and structured phase patterns. This comparative analysis confirms PPO's superiority in both convergence speed and training stability, establishing it as a compelling reinforcement learning approach for optical processor design and implementation.

Following these simulations, we experimentally evaluated the performance of our PPO-based *in situ* reinforcement learning framework on an energy focusing task as shown in **Fig. 3a**. A trainable phase pattern displayed on an SLM modulated the incoming wavefront, and the resulting intensity distribution was recorded on an image sensor plane subdivided into ten detection regions. The objective was to maximize the energy in the designated target region relative to the total energy across all ten regions. This selective beam focusing task provides a benchmark for precisely shaping and controlling light using diffractive optics. **Figure 3b** demonstrates that PPO achieved significantly faster and more effective energy focusing onto target/desired regions. This advantage is visually confirmed in **Fig. 3c**, where the PPO-trained system produced a high-intensity focal spot earlier and with greater clarity. We further conducted experiments with a random, unknown diffuser (**Fig. 4c**) inserted between the SLM and the image sensor plane, as illustrated in **Fig. 4a**. The curves in **Fig. 4b** and the results in **Fig. 4d** further highlight PPO's robustness, demonstrating its ability to maintain effective focusing even in the presence of unknown optical perturbations introduced by a random diffuser. These results provide experimental evidence that our model-free PPO framework enables robust and highly efficient *in situ* optimization of a diffractive optical processor.

We next evaluated our *in situ* reinforcement learning framework on a holographic image generation task, where a phase only SLM was optimized to produce a target image at the sensor plane, as shown in **Fig. 5a**. We experimentally tested the system on two distinct targets: a synthetic grating and a natural image ("Boat"). As shown in **Fig. 5b**, PPO achieved a higher Peak Signal-to-Noise Ratio (PSNR) in less training time. The visual evolution of the grating image generation in **Fig. 5c** further confirms that PPO produced sharper, higher-fidelity images using fewer iterations. This ability to efficiently learn complex optical transformations *in situ* highlights our method's potential for applications such as holographic displays and lensless imaging.

We further extended the PPO-based reinforcement learning framework to an *in situ* image improvement task designed to correct for system aberrations and misalignments. For these experiments, we started with an optical generative model—comprising a digital encoder and an optical decoder (SLM)—that were jointly pre-trained *in silico*[7]. This generative model provided an initial phase pattern to synthesize, from random noise, a novel image following a target/desired distribution; in other words, different from **Fig. 5**, here the image to be projected is a novel image created by the optical generative model. We then used PPO to fine-tune and adaptively optimize the phase pattern of the optical decoder layer directly on the hardware (**Fig. 6a**). This *in situ* reinforcement learning process allows the generative system to learn and compensate for the physical imperfections in the optical decoder hardware. Our experimental results demonstrate a consistent improvement in image quality across various image generation tasks. As quantified in **Fig. 6b**, this PPO-based calibration led to a marked increase in PSNR. The visual results of these experiments, shown in **Fig. 6c,** also illustrate this enhancement. With the *in situ* reinforcement learning, an improvement in image clarity and a more accurate match to the corresponding *in silico*-generated images can be observed. These experimental results further establish our PPO-based reinforcement learning framework as a powerful and efficient method for correcting experimental aberrations and misalignments in dynamic task-specific optical systems.

Finally, we validated the general applicability of our PPO-based *in situ* learning framework for optical computing on an image classification task, specifically classifying handwritten digits (MNIST dataset). As a proof of concept, the experimental setup (**Fig. 7a**) used an 800×800 region of the SLM as a single layer, where the input digit was simultaneously phase-encoded directly onto the same SLM plane. The trainable diffractive layer was optimized *in situ* to learn the class-specific patterns and smartly guide the optical energy toward the detector corresponding to the correct digit class, i.e., performing all-optical image classification. **Figure 7b** plots the experimental test accuracy over *in situ* training epochs. We observed rapid improvement in classification performance during the initial training stages of the reinforcement learning process. Insets show examples of the captured intensity patterns at selected epochs, clearly demonstrating how initially indistinct outputs evolved into clear and discriminative patterns as the *in situ* training progresses. Ten illustrative classification examples for handwritten digits from "0" to "9" are visualized in **Fig. 7c**. Each digit, encoded as a phase input image, was transformed through the learned phase profile. The experimentally captured images reveal clear peaks at the correct class-specific detector positions. The corresponding class scores, derived from these measured intensities, confirm that each digit was correctly identified by the diffractive optical processor. These results underscore the effectiveness of our PPO-based *in situ* learning method for directly training optical processors, offering a robust pathway for developing physical neural networks without reliance on digital twins or model-based error back-propagation.

## Discussion

In this work, we introduced a practical and efficient framework for the *in situ* training of diffractive optical processors using proximal policy optimization. Our model-free PPO-based reinforcement learning method effectively bypasses the gap between simulations and real-world experiments by learning directly from physical measurements. One of the key strengths of our framework is its significant improvement in *in situ* learning speed: by enabling multiple updates from a single batch of experimental data, PPO significantly reduces the number of physical measurements needed for convergence. The clipped surrogate objective used in PPO plays a pivotal role in this acceleration. By preventing large policy updates and enforcing stable improvement, PPO-based reinforcement learning ensures steady and robust convergence even under noisy, limited, or imperfect measurement conditions. This robustness allows for more aggressive optimization without sacrificing stability, further contributing to the speed-up. We validated this improvement through both simulations and experiments. For example, in our optical image classification task, simulations showed that PPO achieved more than 3 × faster convergence compared to other RL methods. Similar advantages in learning speed were also observed in experimental results of targeted energy focusing and holographic image generation. Moreover, we also demonstrated the real-world experimental applicability of PPO-based reinforcement learning for *in situ* aberration corrections and optical image classification tasks, where the PPO-based training enabled robust and efficient system optimization directly from physical feedback.

However, several limitations remain for further improvements. The current policy used Gaussian distributions to sample trainable parameters of the optical system, which did not incorporate prior knowledge about the spatial properties of the physically realizable optical fields or structures. This can lead to inefficient exploration in a high-dimensional space[48]. To address this, future work could explore more sophisticated or expressive policy parameterizations, such as spatially correlated distributions or generative physical priors trained offline to guide the RL-based exploration in a physically conditioned subspace[49–51]. These approaches could substantially reduce the effective dimensionality of the search space, enabling faster learning with fewer iterations. Another promising future direction involves hybrid modeling methods that integrate coarse physical models with our data-driven PPO framework. In this case, an approximate model could guide the early stages of the training process or generate synthetic rollouts that accelerate policy learning. As PPO continues to refine the policy using real-world measurements, the physical model itself could be iteratively updated to better align with the true system dynamics[1,27,52,53]. This closed-loop refinement can allow for improved initialization, reduced data requirements, and a synergistic blend of model-based and model-free learning. Such hybrid strategies are particularly valuable in complex or partially known/characterized physical systems.

From a broader perspective, while our demonstrations have been for optical computing or information processing tasks such as image generation, aberration correction, and optical image classification, the underlying methodology is general and broadly applicable to other experimental learning tasks. Any physical system characterized by a measurable output and adjustable parameters within a feedback loop could potentially benefit from PPO-based *in situ* reinforcement learning. This approach is particularly promising in fields like adaptive optics, nonlinear photonics, multi-mode fiber-optics and other domains where model-free, feedback-driven optimization methods are critical for achieving robust performance in real-world environments[5,54] that are particularly hard to model. Ultimately, this approach moves us toward intelligent, reconfigurable physical systems that can autonomously learn and adapt directly from real-time interactions with their environment.

## Methods

**Optimization Objective**

Our diffractive optical system used an SLM to serve as both the input interface and the trainable optical modulation processor. Let $\varphi \in \mathbb{R}^{H \times W}$ denote the phase modulation pattern applied to the SLM. For a given optical input $x_i$, the system performs a transformation $f(x_i; \varphi)$, which corresponds to the light propagation and modulation governed by the phase mask $\varphi$. The resulting intensity distribution is measured by a digital camera. The goal is to train the system to approximate a desired optical transformation by minimizing the following loss:

$$L(\varphi) = \frac{1}{N} \sum_{i=1}^{N} ||f(x_i; \varphi) - y_i||^2 \qquad (2)$$

This loss is computed directly from physically measured intensities at the sensor plane. Since $f(\cdot)$ represents the actual light propagation and detection process within our experimental setup, the optimization inherently captures all system nonidealities, imperfections, aberrations, misalignments and noise factors.

## Supplementary Information

Supplementary Information file includes:

- Proximal Policy Optimization for In Situ Reinforcement Learning
- Implementation Details

55  Goodman JW. *Introduction to Fourier Optics*. Roberts and Company Publishers, 2005.

**Figures**

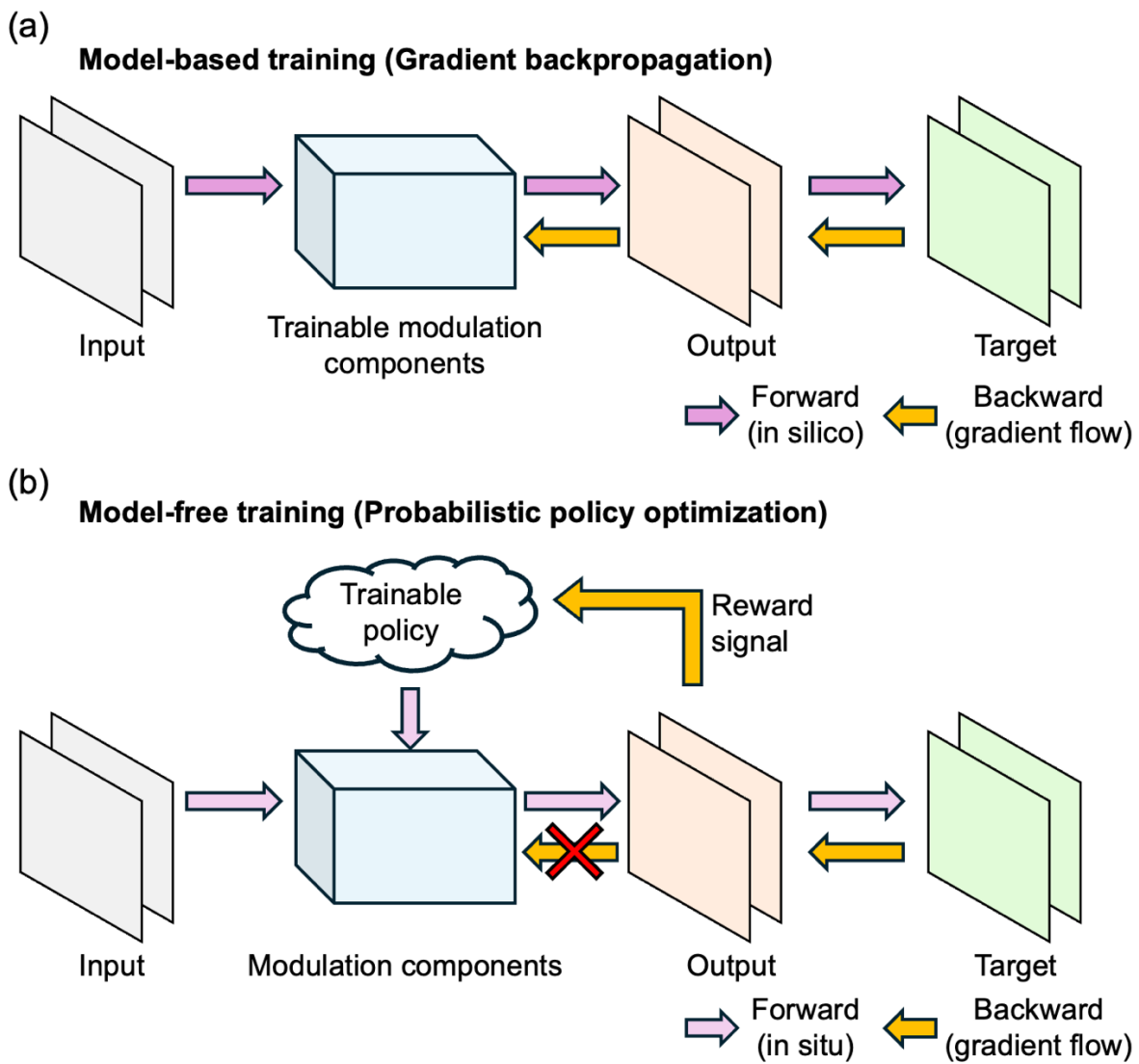

**Figure 1 Conceptual overview of model-based and model-free optical learning**. (a) Model-based training relies on gradient backpropagation through a differentiable *in silico* model of the system. Error gradients are calculated and propagated backward to directly update the system's trainable components. (b) Model-free training employs probabilistic policy optimization to work directly with the physical hardware (*in situ*). A trainable policy generates system parameters, and a reward signal calculated from the measured output is used to update the policy. This reinforcement learning-based method treats the physical system as a black box and does not require a differentiable model or gradient flow through the components.

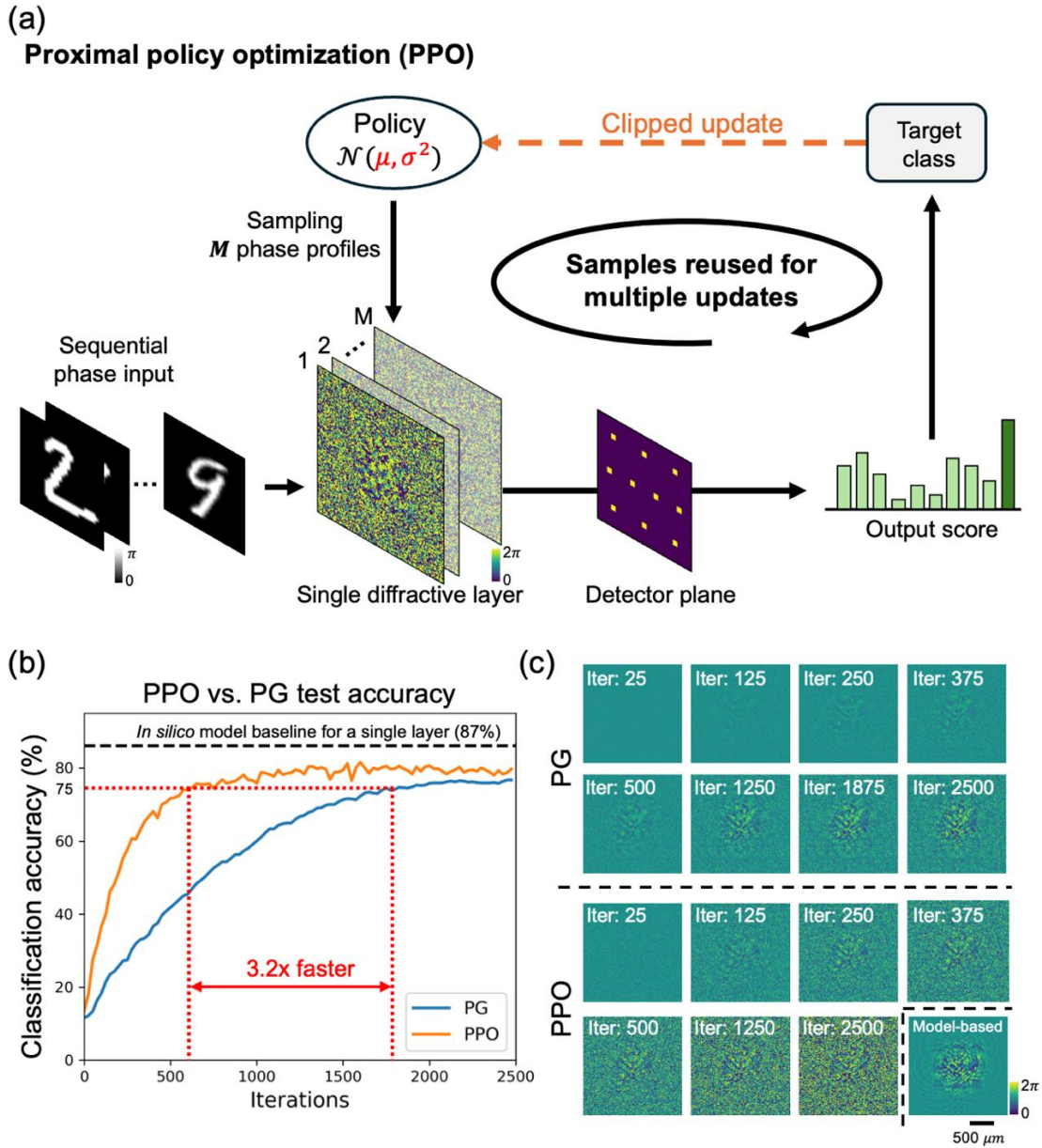

**Figure 2** Proximal Policy Optimization (PPO) for *in situ* training of a diffractive optical classifier. (a) Schematic of the PPO training framework. A policy generates phase patterns for a single diffractive layer. The physical output is measured, and PPO uses a clipped update rule and reuses the measured data for multiple optimization steps to efficiently train the policy. (b) Numerical test accuracy comparison between PPO and standard Policy Gradient (PG). PPO achieves higher classification accuracy and converges faster. The dashed line denotes the performance thresholds, defined as 85% of the *in silico* model performance for a single-layer diffractive configuration. (c) Evolution of the learned diffractive phase patterns at different iterations of PPO and PG compared with the *in silico* model (denoted as Model-based).

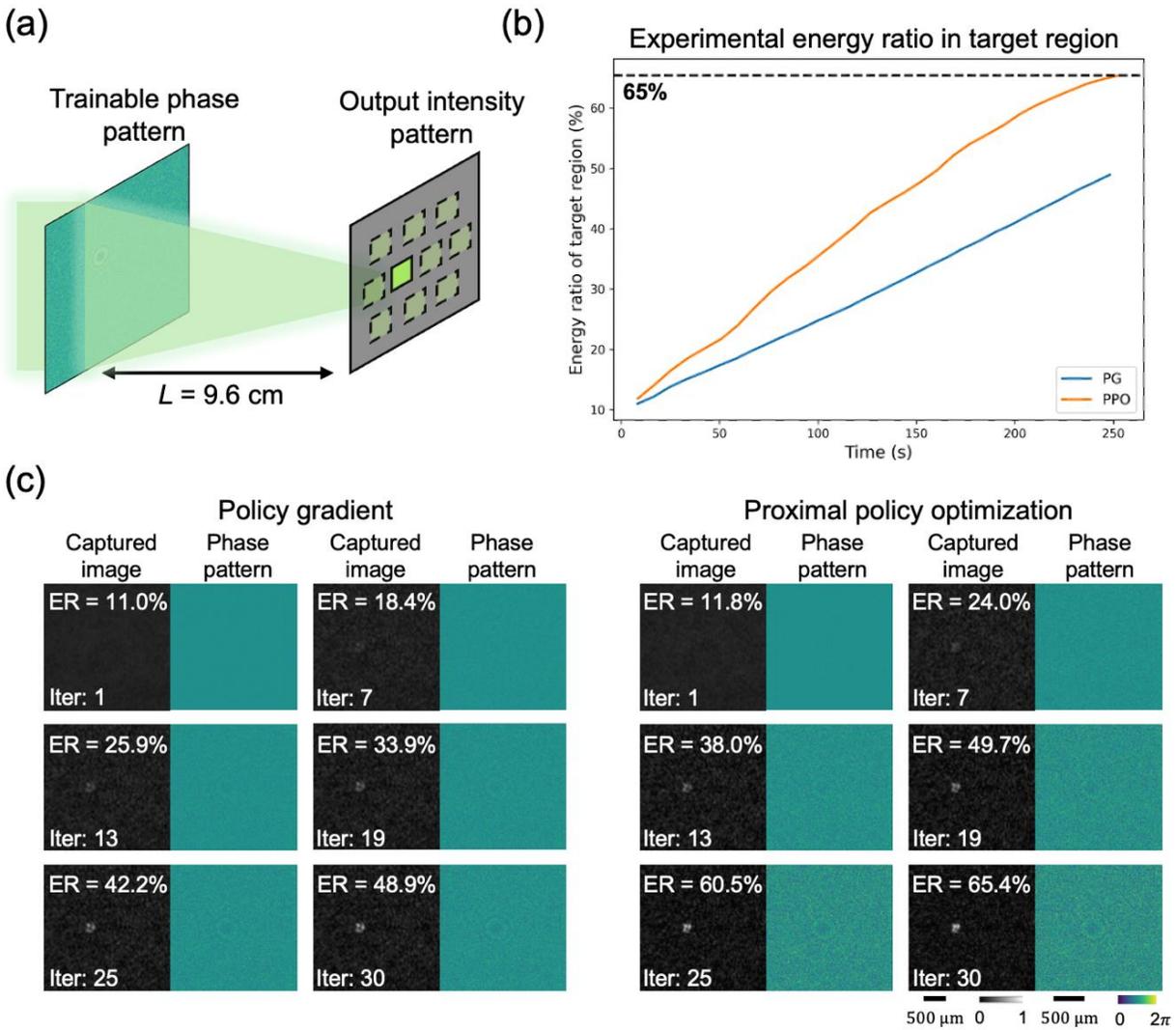

**Figure 3 Experimental reinforcement learning results for *in situ* optimization of targeted energy focusing.** (a) A schematic of the physical setup, wherein a trainable phase pattern is optimized in real-time to focus light onto a designated target area while minimizing energy in the other 9 selected areas. (b) The energy ratio (ER) concentrated in the target/desired region is plotted as a function of the experimental reinforcement learning time. (c) Time-lapse visualization of the captured intensity patterns and the corresponding SLM phase patterns for PG (left) and PPO (right) during the optimization.

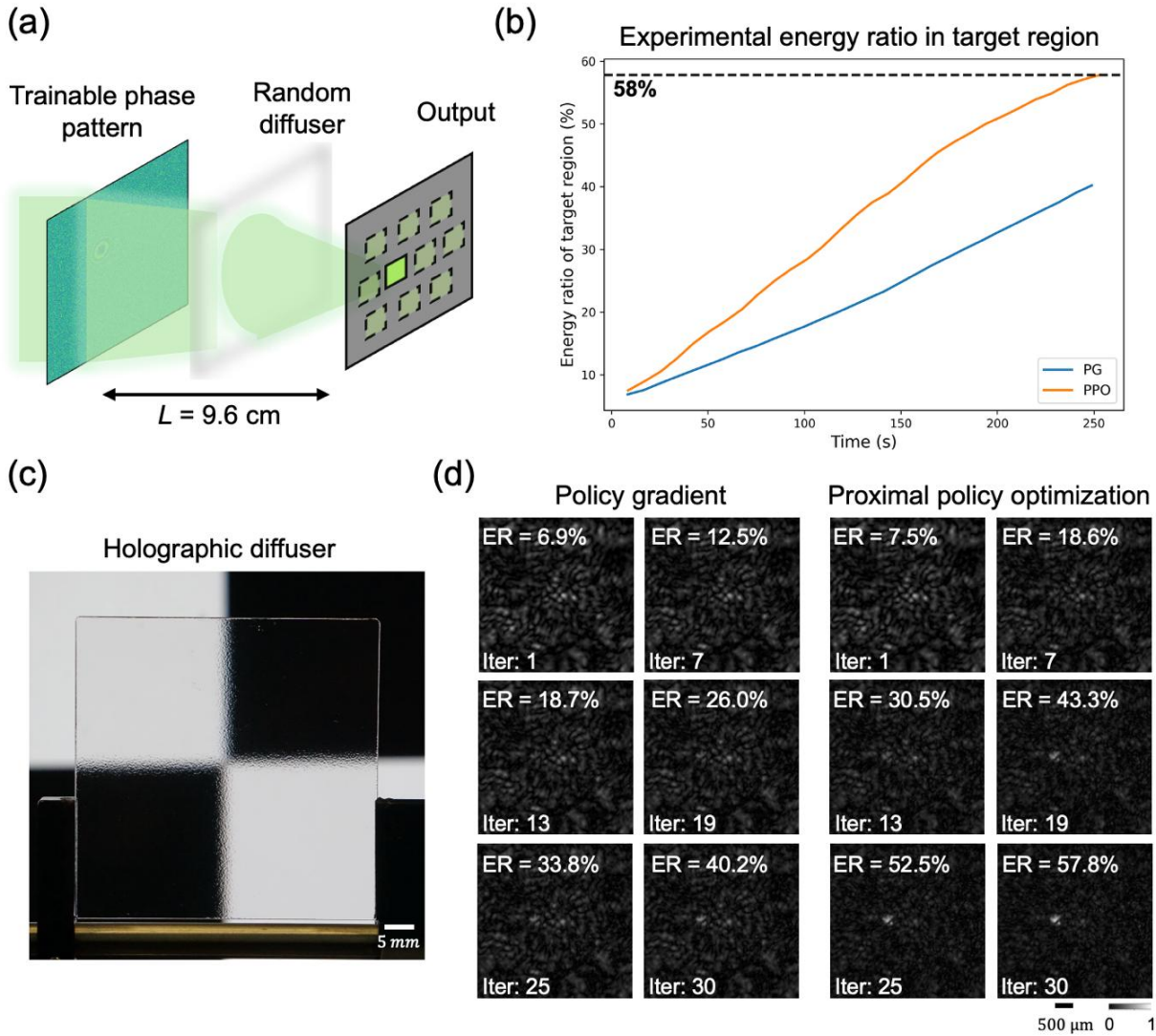

**Figure 4 Experimental reinforcement learning results for *in situ* optimization of targeted energy focusing with a random, unknown diffuser inserted between the SLM and the image sensor plane.** (a) A schematic of the physical setup, wherein a trainable phase pattern is optimized in real-time to focus light through an unknown, random diffuser onto a designated target area while minimizing energy in the other 9 selected areas. (b) The energy ratio (ER) concentrated in the target region is plotted as a function of experimental reinforcement learning time. (c) Photograph of the unknown, random diffuser used in the experiment. (d) Time-lapse evolution of the captured intensity patterns for PG (left) and PPO (right) during the optimization with the random diffuser inserted.

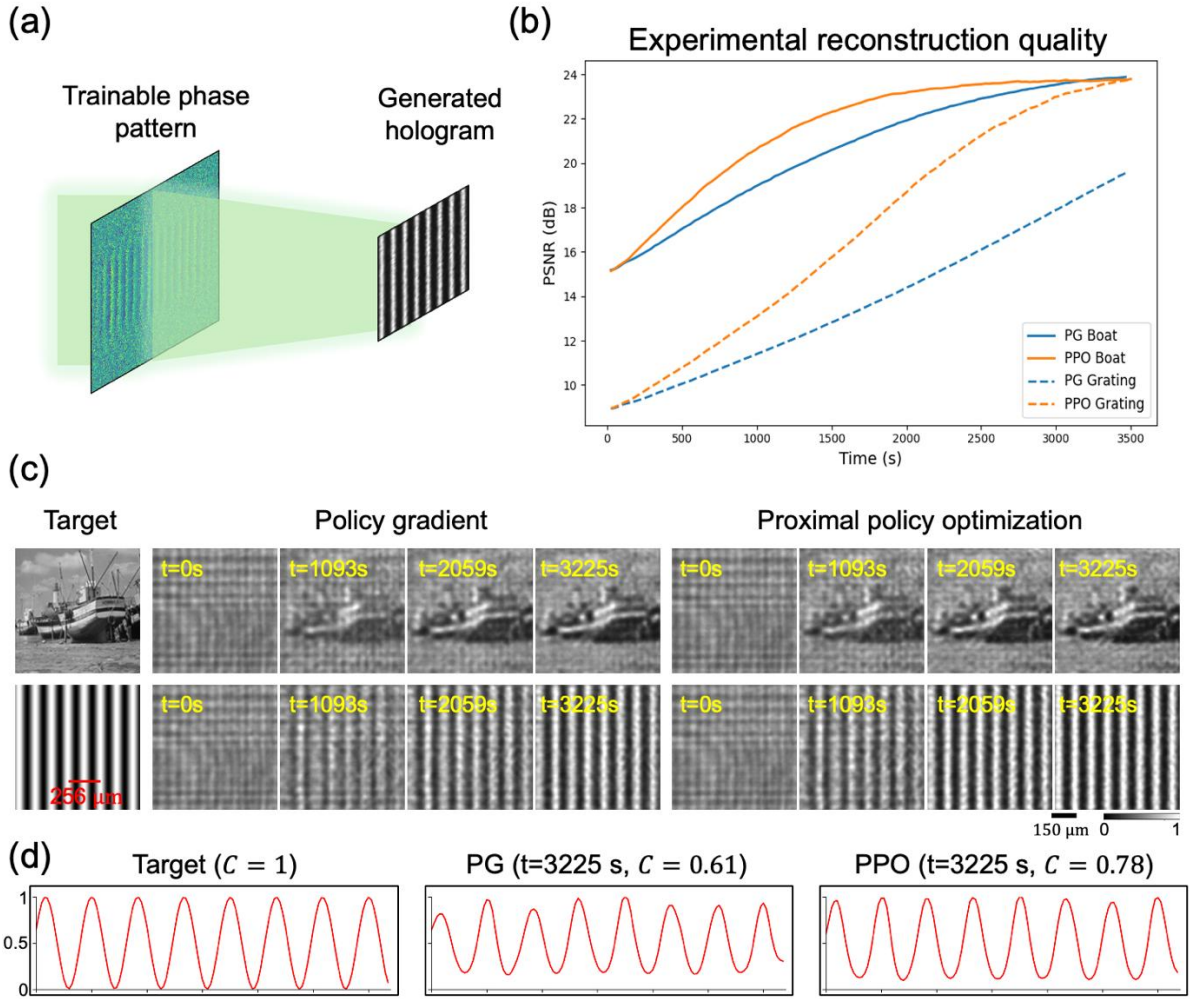

**Figure 5 Experimental reinforcement learning results for *in situ* holographic image generation.** (a) Schematic of the experimental setup for holographic image projection, where a trainable phase pattern generates a selected target image (the initial condition at t = 0 s is a uniform phase pattern). (b) Reconstruction quality (PSNR) over time for two target images, a 'Boat' and a 'Grating'. PPO (shown with orange lines) consistently achieves higher fidelity faster than PG (blue lines). (c) Visual comparison of holographic image quality at different time points. PPO reconstructions (right) are clearer and more accurate than PG reconstructions (middle) at all stages. (d) A cross-sectional intensity profiles of the 'Grating' pattern. Contrast values (C) were calculated, showing that PPO achieves a higher contrast.

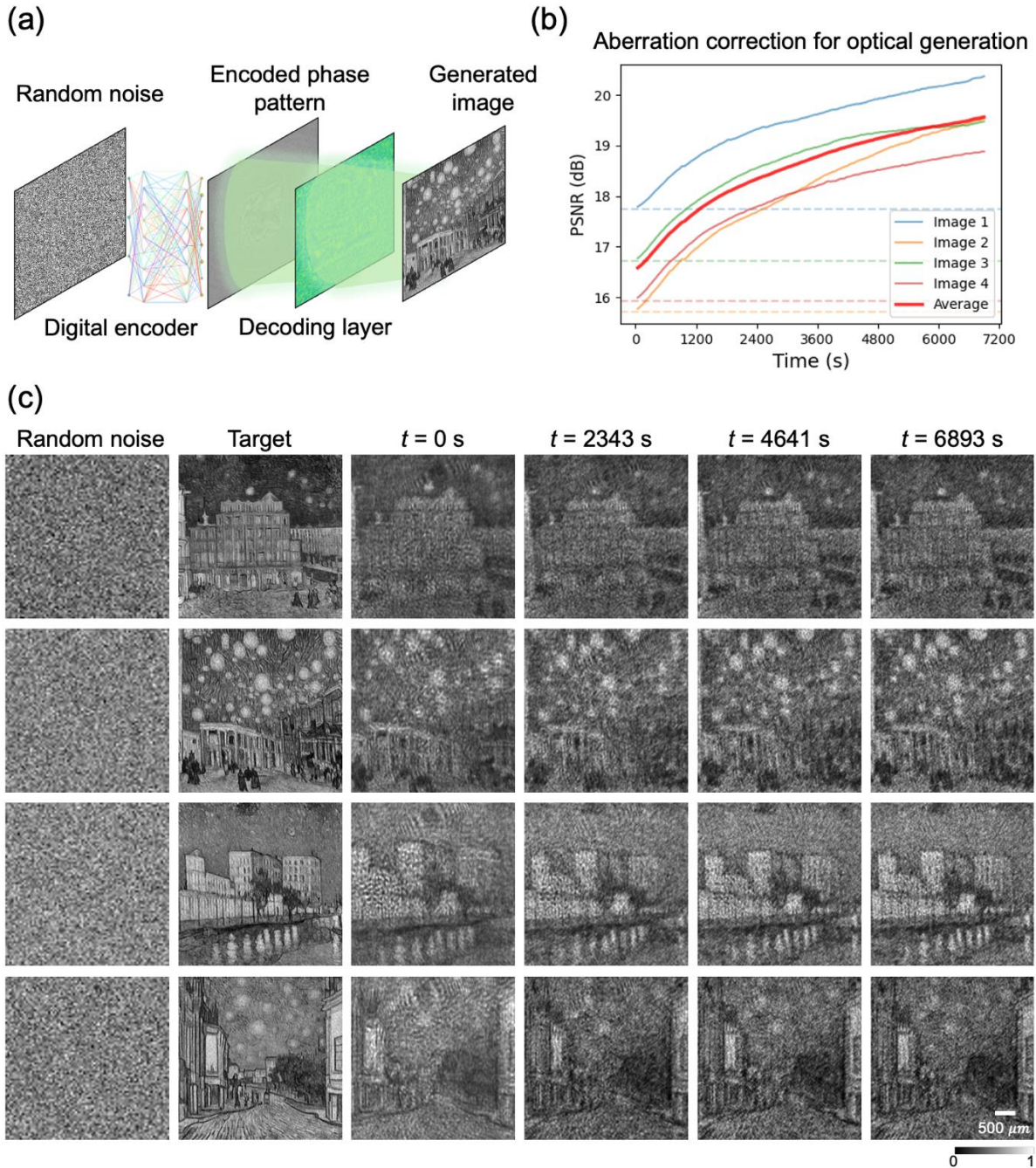

**Figure 6** *In situ* **aberration correction for synthesizing novel images through an optical generative model.** (a) Schematic of the optical generative model, where a digital encoder transforms random noise into an encoded phase pattern, which is displayed on an SLM. This encoded phase pattern then passes through a decoding layer (the aberrated optical system) to form a novel image; different than Fig. 5, here the target images are novel, synthesized by the optical generative model. (b) Reconstruction quality (PSNR) as a function of the *in situ* training time for four different novel images created from random noise. The average performance (red line) shows a consistent improvement, demonstrating successful *in situ* aberration correction through reinforcement learning. (c) Visual results of the RL-based optimization process. The generated novel images from random input noise, initially aberrated at t=0s, become progressively clearer and more recognizable over time as the system learns *in situ* to compensate for aberrations.

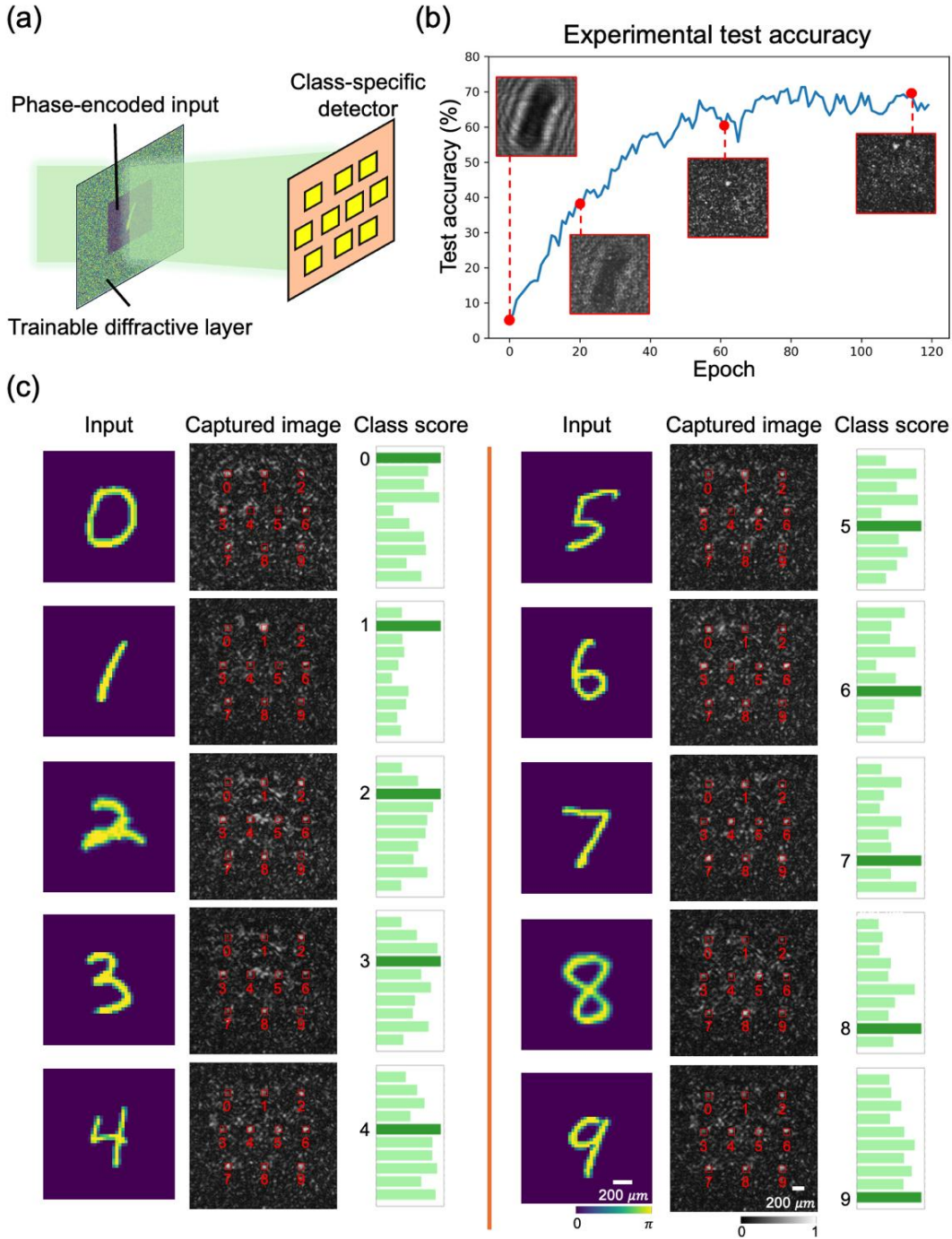

**Figure 7 Experimental demonstration of *in situ* reinforcement learning for an all-optical diffractive image classifier.** (a) A schematic of the optical setup where a trainable diffractive layer

processes phase-encoded input images (to be classified and never seen before) and directs the input light to a grid of 10 class-specific detectors used for all-optical image classification. (b) Experimental test accuracy plotted against the *in situ* training epoch number, showing the learning progression of the physical system. Insets visualize the evolution of the output intensity patterns at various stages of the RL process; once converged, the learned phase pattern is fixed and can be used to classify new test objects never seen before. (c) Classification test examples for ten different handwritten digits {0, 1, ..., 9}, never seen before. The final learned phase profile successfully directs the light from each input digit to its correct corresponding detector, as confirmed by the captured images and class scores.